\documentclass[fleqn,11pt,oneside]{article}

\usepackage{booktabs} 
\usepackage{tabularx}
\usepackage{orcidlink}  
\usepackage{setspace} 
\usepackage{bbm}  
\usepackage{bm}  
\usepackage[normalem]{ulem} 
\usepackage{amsmath,amsthm,amssymb,amscd} 
\usepackage{natbib} 
\usepackage[top=1.1in, bottom=1.1in, left=1.2in, right=1.1in]{geometry} 
\usepackage{caption} 
\usepackage{hyperref} 
\usepackage{chngcntr} 
\usepackage{todonotes}
\usepackage{authblk}

\usepackage[linesnumbered,ruled,vlined]{algorithm2e}
\usepackage{algpseudocode}

\SetKwInput{KwInput}{Input}    
\SetKwInput{KwOutput}{Output}

\renewcommand{\cite}{\citep}

\newcommand{\td}[1][]{\textcolor{black}{\fcolorbox{black}{lightgray}{%
\textbf{\textcolor{black}{\scalebox{0.7}[1.0]{\small TODO}}}}%
\ifthenelse{\equal{#1}{}}{}{~\emph{#1}}}\xspace}

\usepackage[ruled]{algorithm2e}

\SetAlFnt{\small}
\SetAlCapFnt{\small}
\SetAlCapNameFnt{\small}
\SetAlCapHSkip{0pt}

\def\1{\mathbbm{1}}

\begin{document}
\title{\vspace*{-6ex}Closed-Form Interpretation of Neural Network Latent Spaces with Symbolic Gradients\vspace*{-1ex}}

\author[1,2,3]{Sebastian J. Wetzel}
\author[4,5]{Zakaria Patel}

\affil[1]{University of Waterloo, Waterloo, Ontario N2L 3G1, Canada}

\affil[2]{Perimeter Institute for Theoretical Physics, Waterloo, Ontario N2L 2Y5, Canada}
\affil[3]{Homes Plus Magazine Inc., Waterloo, Ontario N2V 2B1, Canada\\
  \texttt{swetzel@perimeterinstitute.ca}}
\affil[4]{Stability AI Ltd., London, UK}
\affil[5]{University of Toronto, Toronto, Ontario M5S 1A1, Canada \\
  \texttt{zakaria.patel@stability.ai}}

\date{\today\vspace*{-4ex}}

\date{}
\maketitle

\begin{abstract}
It has been demonstrated that artificial neural networks like autoencoders or Siamese networks encode meaningful concepts in their latent spaces.
However, there does not exist a comprehensive framework for retrieving this information in a human-readable form without prior knowledge. In quantitative disciplines concepts are typically formulated as equations. 
Hence, in order to extract these concepts, we introduce a framework for finding closed-form interpretations of neurons in latent spaces of artificial neural networks.
The interpretation framework is based on embedding trained neural networks into an equivalence class of functions that encode the same concept. We interpret these neural networks by finding an intersection between the equivalence class and human-readable equations defined by a symbolic search space. Computationally, this framework is based on finding a symbolic expression whose normalized gradients match the normalized gradients of a specific neuron with respect to the input variables.
The effectiveness of our approach is demonstrated by retrieving invariants of matrices and conserved quantities of dynamical systems from latent spaces of Siamese neural networks.
\end{abstract}

\vspace{2pc}
\newpage

\section{Introduction}

\label{sec:intro}

The current AI revolution is driven by artificial neural networks (ANNs). These models have enabled machines to achieve superhuman performance in a variety of tasks, such as image recognition, language translation, game playing, and even generating human-like text. However, this remarkable power comes at the expense of interpretability, often referred to as the black-box problem. The representational capacity of artificial neural networks relies on interactions between possibly billions of neurons. While each single neuron is easy to describe mathematically, as networks become larger, it becomes increasingly difficult to understand how these interactions give rise to a neural network's overall prediction. 

The black-box nature of neural networks can be acceptable in applications where prediction is the primary goal. However, in science, where the goal is not just prediction but also understanding the underlying phenomena, interpretability is crucial. Moreover, in medicine, it is important to understand why an AI system has made a particular diagnosis or treatment recommendation to avoid risks of dangerous or ethically questionable decisions \cite{Jin2022,Amann2022}. AI interpretability in the law domain is crucial for understanding and explaining how automated decisions are made, which helps ensure transparency and accountability. It also allows for the identification and correction of biases, compliance with regulations, and maintains the integrity of legal processes \cite{Hacker2020,Bibal2020}. 

In many scientific applications of neural networks, it can be verified that neural networks often learn meaningful concepts, similar to those that humans use, to describe certain phenomena \cite{Ha2021,Desai2021,PoulsenNautrup2022} . Unfortunately, without a method to distill this learned concept in a human-interpretable form, the only way to reveal it is by directly comparing it to a set of candidates that the researcher is already aware of. Clearly, it is not possible to make new discoveries in this way. 


To address this problem, symbolic regression techniques have been proposed to interpret neural networks by deriving closed-form expressions that represent the underlying concepts learned by these networks \cite{cranmer2020discovering,Mengel2023}. These approaches involve exploring the space of potential mathematical expressions to identify those that best replicate the predictions of a neural network. Unfortunately, such methods are limited to interpreting output neurons of neural networks performing regression, where the concept that is recovered is the global function learned by the neural network. 

Neural networks applied to perform scientific discovery are often tasked with solving problems that cannot be formulated under the umbrella of regression. Moreover, often relevant information is encoded by neurons deep inside neural networks, specifically in latent spaces. Prominent artificial scientific discovery methods have been proposed based on networks like autoencoders \cite{Wetzel2017,Iten2020,Miles2021b,Frohnert2024} or Siamese networks \cite{Wetzel2020,Patel2022,Han2023}. These networks can distill meaningful concepts inside their latent spaces without being supplied with explicit training information in the form of labeled targets. The crucial obstacle to their wider adoption is the lack of tools that enable the recovery of the learned concepts without prior knowledge. Removing this bottleneck would allow scientists to use these tools to discover potentially new scientific insights.

For these reasons, it is desirable to have a framework capable of interpreting concepts encoded in arbitrary intermediate neurons of artificial neural networks. In this paper, we describe such a framework that can be employed to interpret any single neuron within an artificial neural network in closed form.

The central barrier that we need to overcome is that concepts encoded in neurons in hidden layers are generally not stored in a human-readable form, but instead get distorted and transformed in a highly non-linear fashion, which limits the naive employment of symbolic regression techniques. The three obstacles towards discovering a closed-form functional interpretation of information encoded inside neural networks are:
\begin{enumerate}
\item \textbf{scaling of symbolic representations:} Any form of symbolic search algorithm scales poorly with the complexity of the underlying equation. Many scientists are working on competing symbolic search algorithms mainly tailored to symbolic regression, a list can be found in the subsequent paragraph.
\item \textbf{dimensional mismatch} of neural networks storing information distributed among multiple neurons. Common methods to eliminate this mismatch are based on disentangling features learned by different neurons within the same layer \cite{higgins2017beta} or to enforce a bottleneck \cite{pmlr-v119-koh20a} such that single neurons capture individual concepts.
\item \textbf{distortions of concepts} within a neural network in a highly non-linear form. If neural networks learn concepts, there is no reason to store them in a form aligned with a human formulation of the concept. For example, if a neural network learns the concept of temperature, there is no reason to choose the Celsius or the Fahrenheit scale, nor does this encoding need to be linear. In practice, it turns out that this non-linear distortion cannot even be captured with symbolic equations. This problem prevents symbolic search algorithms from interpreting anything beyond output neurons in the context of regression. Until the invention of the interpretation framework presented in our manuscript, solving this problem was impossible.
\end{enumerate}
The interpretation method presented in this manuscript is currently the only option to overcome obstacle 3 and is highly complementary with other publications that overcome obstacles 1 and 2. 

Our proposed solution is based on the idea of retrieving information encoded inside a desired neuron instead of replicating the neuron's output symbolically. The interpretation method is based on constructing an equivalence class around this neuron, which contains all functions encoding the same concept as the target neuron. In practice, we interpret the neuron by searching for a closed-form representative function contained in this equivalence class.

This article is structured in the following manner: After reviewing related work, we explain the mathematical basis underpinning the interpretation procedure centered around the equivalance class of all functions containing the same information as a target neuron. Further, we design an algorithm to find the simplest symbolic representative of this equivalence class. We demonstrate the power of our framework by rediscovering the explicit formulas of matrix invariants and conserved quantities from the latent spaces of Siamese networks.


\section{Related Work}
\label{sec:related-work}

The current manuscript concerns the domain of artificial neural network interpretability, with a focus on enabling new scientific discovery through latent space models. Much of the neural network interpretability research adresses the question of whether or not neural networks learn certain known scientific concepts. While verifying a neural network is an important task, it is unsuitable for gaining novel scientific insights. There has been limited progress toward revealing scientific insights in symbolic form from artificial neural networks that do not require previous knowledge of the underlying concept beforehand \cite{Wetzel2017b,cranmer2020discovering,Miles2021b,Liu2021}. These cases are rare examples where the underlying concept is encoded in a linear manner, or where other properties of the concept simplify the interpretation problem.
While there are no unified approaches to interpreting latent space models, it might in principle be possible to build such models based on architectures with symbolic layers \cite{Martius2016,sahoo2018learning,Dugan2020,https://doi.org/10.48550/arxiv.2404.19756}

Our article aims instead to interpret existing latent space models. We extend an interpretation framework \cite{https://doi.org/10.48550/arxiv.2401.04978}, originally developed to interpret neural network classifiers, to interpret neural network latent spaces.

The interpretation method relies on efficiently searching the space of symbolic equations, which can be achieved by genetic search algorithms which form the backend of many symbolic regression algorithms. These include Eureqa \cite{Schmidt2009},  Operon C++ \cite{Burlacu2020}, PySINDy \cite{Kaptanoglu2022}, Feyn \cite{Brolos2021}, Gene-pool Optimal Mixing Evolutionary Algorithm \cite{Virgolin2021}, GPLearn \cite{Stephens2022} and PySR \cite{Cranmer2023}. Other symbolic regression algorithms include deep symbolic regression uses recurrent neural networks \cite{petersen2020deep}, symbolic regression with transformers \cite{kamienny2022end,pmlr-v139-biggio21a} or AI Feynman \cite{Udrescu2020}. 

An overview of interpretable scientific discovery with symbolic Regression can be found in \cite{Makke2022,Angelis2023}.

\begin{figure}[!h]
    \centering
    \includegraphics[width=\textwidth]{method_comparison.pdf}
    \caption{(a) The Siamese network consists of two pairs of identical sub-networks $f$. From the first pair, we compute the distance between the anchor and the positive example $d(f(x_A), f(x_P))$, which should be as close to zero as possible. From the second we compute $d(f(x_A), f(x_N))$, which should be as large as possible. This facilitates a latent space where similar items are close together, while dissimlar ones are far apart. (b) Most existing approaches attempt to interpret a neural network latent space by comparing the latent with known candidate concepts. In this case, it is necessary to have the correct concept at hand, which is unsustainable for scientific discovery. (c) Our method requires only a dataset and a trained neural network to be used in conjunction with a symbolic search algorithm, which then discovers a closed-form expression describing the concept encoded in the network's latent space.}
    \label{fig:method_comparison}
\end{figure}

\section{Method}
\label{sec:method}
\subsection{Siamese Neural Networks}
\label{sec:siamese_neural_networks}
Siamese neural networks (SNN) \cite{Baldi1993,bromley1993signature} were originally introduced to solve fingerprint recognition and signature verification problems. SNNs consist of two identical sub-networks with shared parameters, each receiving distinct inputs which are then projected to an embedding space. These projections are then compared by a distance metric, which joins each sub-network $f$ together at their output. Inputs belonging to the same class should obtain high similarity, while those belonging to different classes should obtain low similarity. Such a framework allows for generalization to infinite-class classification problems. The distance metric $d(\cdot)$ is chosen according to the specific problem at hand, and in our case we use the squared Euclidean distance. 

The network $F$ can be trained effectively using a contrastive or triplet loss \cite{tripletLoss}, wherein a set of triplets are supplied to the energy function,
\begin{align*}
    \mathcal{L}(x_A, x_P, x_N) =  \max(d(f(x_A), f(x_P)) - d(f(x_A), f(x_N)) + \alpha, 0).
    \label{eq:siamese}
\end{align*}
The anchor $x_A$ is the ground truth class, the positive sample $x_P$ is of the same class as $x_A$, whereas the negative sample $x_N$ is of a different class. Instead of using a twin network, this setup requires a triplet of identical networks, each still sharing the same weights. The triplet loss is minimized when the distance between the anchor and positive sample is minimized in the embedding space, while the distance between the anchor and negative sample is maximized. The margin parameter $\alpha$ is a positive constant which encourages separation between positive and negative samples, as $\alpha=0$ would mean that the loss could be trivially minimized by projecting all samples to the same location. Finally, the $\max(\cdot)$ operation ensures that the distance between positive and negative samples remains finite. 

It has been shown that in scientific settings SNNs can be trained to learn conserved quantities and symmetry invariants of the underlying system. For this purpose, training data is collected where data points belonging to the same class are defined through a connection via trajectories obeying laws of motion (conserved quantities) or a desired symmetry group (symmetry invariants) \cite{Wetzel2020}.

The architecture of the sub-network $f$ depends on the underlying data. In our case, we implement it as a fully-connected network. We note that our framework interprets single neurons, hence our latent layer (i.e., the final neuron in our sub-network $f$), which we wish to interpret, consists of only one neuron. The details of our architecture and training hyperparameters can be found in \autoref{appdx:neural_network}.


\subsection{Interpretation Framework}
\label{sec:framework}


The interpretation framework is designed to extract concepts in the form of symbolic equations from any single disentangled or concept bottleneck neuron within an artificial neural network. While the interpretation framework can be applied to any single neuron, for the purpose of this manuscript we perform an interpretation of the output neuron $f(\mathbf{x})$ of a single sub-net of a Siamese network defined by \eqref{eq:siamese} which produces a scalar mapping of the input into a latent space.

$f(\mathbf{x})$ contains the full information about a certain symbolic concept $g(\mathbf{x})$ if $g(\mathbf{x})$ can be faithfully reconstructed from $f(\mathbf{x})$. Conversely, if $f(\mathbf{x})$ only contains information from $g(\mathbf{x})$ it is possible to reconstruct $f(\mathbf{x})$ from the knowledge of $g(\mathbf{x})$. In mathematical terms that means that there exists an invertible function $\phi$ such that $f(\mathbf{x}) = \phi(g(\mathbf{x}))$. An example of the same concept embedded in different forms is the temperature, it can be measured in Fahrenheit or Celsius and there exists a linear transformation that maps one version of the temperature onto the other.

In general, this means that if we aim to extract information from a neural network $f$, we need to account for any nonlinear and uninterpretable transformation $\phi$ that conceals the human formulation of a concept,
\begin{align}
f(\mathbf{x})=\underbrace{\phi}_{\text{uninterpretable transformation}} ( \underbrace{g(\mathbf{x})}_{\text{closed form concept} } ). 
\end{align}
Different realizations of neural networks might learn the same concept $g$ and therefore contain the same information.. Each of these networks might deform the concept by a different $\phi$. More formally, these realizations are all members of the following equivalence class:
\begin{align}
\label{eq:equivalence_class_1}
\widetilde{H}_g = \left\{ f(\mathbf{x}) \in C^1(D \subset \mathbb{R}^n, \mathbb{R}) \mid \exists \text{ invertible } \phi \in C^1(\mathbb{R}, \mathbb{R}) : f(\mathbf{x}) = \phi(g(\mathbf{x})) \right\}.
\end{align}
While each network $f\in\widetilde H_g$ is related to $g$ via a different unique invertible transformation $\phi$, they are functionally equivalent in that they learn the same concept from the data. At this point, we ask the question, whether it is possible to identify the concept $g$ without knowing the function $\phi$.
%
%
\begin{align}
    g(\mathbf{x}) = \phi^{-1} \left(f(\mathbf{x})\right).
\end{align}
%
%
In order to avoid the necessity of knowing $\phi$, we rewrite the equivalence class \eqref{eq:equivalence_class_1} such that membership can be defined without explicit information about $\phi$. Since all $f\in \widetilde H_g$ are required to be continuously differentiable, we can show that the gradients of the two functions $f$ and $g$ point in the same direction,
\begin{align}
    \nabla f(\mathbf{x}) = \phi'(g(\mathbf{x})) \cdot \nabla g(\mathbf{x}) \quad \text{where} \quad \|\phi'(g(\mathbf{x}))\| > 0. 
\end{align}
Here we used that $\phi$, by construction, is invertible, which means $\phi$ is monotonoous. Since $\phi'(g(\mathbf{x}))$ is merely a scaling factor, this equation allows us to define a new equivalence class $\widetilde H_g \subseteq H_g=H_{g+}\cup H_{g-} $, where 

\begin{align}
    \label{equivalence_class}
    H_{g\pm} = \left\{f(\mathbf{x}) \in C^ {1} (D \subset \mathbb{R}^n, \mathbb{R}) | \frac {\nabla f(\mathbf{x})}{\left \| \nabla f(\mathbf{x})\right\|} = \frac {\pm\nabla g(\mathbf{x})}{\left\| \nabla g(\mathbf{x})\right\|} \vee \nabla f(\mathbf{x})=\nabla g(\mathbf{x})=0, \forall \mathbf{x} \in D \right\}.
\end{align}
Trivially, if $f\in H_g$ then $H_g=H_f$. It can be proven that $H_g = \widetilde{H}_g$, see \autoref{sec:proof} under mild assumptions. In \autoref{sec:assumptions} we explore whether these assumptions are justified in typical neural network settings. In order to execute the interpretation framework we look at the definition of this equivalence class in reverse. We define an equivalence class anchored on the neural network $H_f$ and use a genetic algorithm to retrieve the most likely symbolic concept $g$ within $H_f$.

\subsection{Symbolic Search}
\label{sec:symbolic_search}
Symbolic search is primarily used in symbolic regression for finding closed-form expressions that approximate the relation between target and input variables for a given dataset. Typically, this is done by employing a genetic algorithm, which evolves a population of candidate formulas using genetic operations like selection, crossover, and mutation, aiming to find the least complex tree of operators $T$ that best maps inputs $X$ to outputs $Y$ according to some objective function. This tree consists of a set of nodes, each containing either a number, variable, or a unary or binary operator (see \autoref{fig:method_comparison} (c) for an example tree) that represent a mathematical expression. In the context of neural network interpretation, symbolic regression is employed to convert a complex model into an interpretable tree representation.

In our case, we search for a symbolic tree $T$ which represents a function $g \in H_{f+}$, meaning, we look for a symbolic concept $g$ within the equivalence class anchored on the neural network $f$. During this step we choose a symbolic quantity whose gradient points in the same direction as the gradient of the network $f$. Focussing solely on $H_{f+}$ is possible because $H_{f-}$ can be mapped to $H_{f+}$ simply by multiplying each element with $-1$. In contrast to symbolic regression, instead of performing regression on a set of prediction targets to find the best fitting function, we search for an analytical expression whose normalized gradients are as close as possible to those of $f$. Because of this difference, we refer to this approach as symbolic search instead of symbolic regression. Note that this requires that $T$ consists of operators that yield a differentiable function. To implement our symbolic search algorithm, we modify the SymbolicRegression.jl module, the backend of the PySR package \cite{Cranmer2023}.

The objective function we choose is the mean-squared-error (MSE), which measures the distance between the normalized gradients $g_T(\mathbf{x}) = \frac{\nabla T(\mathbf{x})} {\|\nabla T(\mathbf{x}) \|}$, and $g_f(\mathbf{x}) = \frac{\nabla f(\mathbf{x})} {\|\nabla f(\mathbf{x}) \|}$, 
\begin{align}
    \text{MSE}(g_T(X), g_f(X)) = \frac{1}{n} \sum_{i=1}^n \left\| g_T(\mathbf{x}_i) - g_f(\mathbf{x}_i) \right\|^2.
\end{align}
Nodes are mutated and added by the modified symbolic search algorithm in order to minimize this objective function. The unary operators we use include $\{\text{sqrt(square root)}, \text{sq(square)}, \text{sin}, \text{exp}\}$, and for binary operators we use $\{+, -, *, /, \wedge \}$. The setup we use is described in \autoref{appdx:sr_params}.


\subsection{Algorithms}

Implementing our framework involves three main algorithms which summarize the preceding sections: 
\begin{enumerate}
    \item Train the model $f_\theta$ to learn the invariant. See \autoref{alg:train_network}.
    \item Choose a neuron to interpret. In our specific case, we are interested in interpreting the latent space of a Siamese network $f_\theta$ with weights $\theta$. Compute its gradient with respect to the input, i.e., \(\nabla_{\mathbf{x}} f_{\theta}(\mathbf{x})\). See \autoref{alg:extract_grad}.
    \item Apply symbolic search to find a symbolic tree $T$ whose gradients point in the same direction as $f_\theta$. See \autoref{alg:symbolic_search}.
\end{enumerate}

\begin{algorithm}[!ht]
    \footnotesize
    \SetAlgoNoLine
    \KwData{Dataset of triplets $\mathcal{D} = \{(X_A, X_P, X_N)_i\}_{i=1}^m $}
    \KwIn{Neural network hyperparameters}
    \KwOut{Trained network $f_\theta$}
    \For{each epoch}{
        \For{each mini-batch $\{(X_A, X_P, X_N)\}$ from $\mathcal{D}$}{
            $f_A = f_\theta(X_A)$ \\
            $f_P = f_\theta(X_P)$ \\
            $f_N = f_\theta(X_N)$ \\            
            $\mathcal{L} = \max(0, \|f_A - f_P\|_2^2 - \|f_A - f_N\|_2^2 + \alpha)$ \\

            Backpropagate the loss and update the model parameters $\theta$
        }
    }
    \caption{Training a Siamese Neural Network to Learn an Invariant}
    \label{alg:train_network}
\end{algorithm}

\begin{algorithm}[!ht]
    \footnotesize
    \SetAlgoNoLine
    \KwData{Unlabelled dataset $(X)$}
    \KwIn{Trained network $f_\theta$}
    \KwOut{$(X, g_f)$}
    $g_f \leftarrow[ \nabla f_{\theta}(\mathbf{x}) \text{ for } \mathbf{x} \text{ in } X]$ \Comment{Evaluate gradients w.r.t. input at neuron $f$} \\
    $g_f \leftarrow[ \frac{\nabla f_{\theta}}{\|\nabla f_{\theta}\| + \epsilon} \text{ for } \nabla f_{\theta} \text{ in } g_f]$  \Comment{Normalize Gradients}
    
    \caption{Extracting the Gradients from the Siamese Network}
    \label{alg:extract_grad}
\end{algorithm}

\begin{algorithm}[!ht]
    \footnotesize
    \SetAlgoNoLine
    \KwData{Gradient data set $(X, g_f)$}
    \KwIn{Symbolic search hyperparameters; a set of unary and binary operations.}
    \KwOut{Symbolic model $T$}
    Initialize symbolic model $T$ \\
    Evolve $T$ \textbf{with} ( \\
    \phantom{aaa} $g_T\leftarrow[ \nabla T(\mathbf{x}) \text{ for } \mathbf{x} \text{ in } X]$ \Comment{Gradients of symbolic model} \\
    \phantom{aaa} $g_T\leftarrow  [\textbf{if }\nabla T(\mathbf{x})\neq 0: \nabla T(\mathbf{x})/ \lVert \nabla T(\mathbf{x}) \rVert $ \\
    \phantom{aaaaaaaaa}$\textbf{ else }\nabla T(\mathbf{x}) \text{ for } \nabla T(\mathbf{x}) \text{ in } g_T$]  \Comment{Normalize Gradients} \\
    \phantom{aaa} ) to minimize MSE$(g_f, g_T)$

    \caption{Symbolic Search}
    \label{alg:symbolic_search}
\end{algorithm}


\section{Experiments}
\subsection{Dataset Generation}
 To test the effectiveness of our method, we demonstrate it on 12 different datasets. Each dataset consists of $50000$ training triplets, $5000$ validation triplets and $10000$ test triplets. The triplets are constructed in the following way: once the anchor $x_A$ is sampled, the positive sample $x_P$ is obtained via $x_P=M(x_A)$, where $M$ is a placeholder operator for a specific transformation that is defined for each experiment in \autoref{appdx:dataset}, and finally $x_N$ is sampled independently. The operation implemented by $M$ transforms $x_A$ to $x_P$ such that certain properties of $x_A$ are inherited by $x_P$, but the two points are otherwise unique. The invariants and conserved quantities corresponding to our data sets are: trace, determinant, sum of principal minors under the similarity transformation, the inner product and spacetime interval under the Lorentz transformation, and the energy and momentum in a variety of potentials. More details about each dataset, including how to reproduce them, can be found in \autoref{appdx:dataset}.

\begin{table}[htbp]
\caption{Matrix Invariants}\label{table:tab_matrix}
\resizebox{\columnwidth}{!}{\begin{tabular}{|l|l|l|l|l|l|l|}
\hline
Exp. No. & Name & $d$ & Transformation & Invariant & Analytical Expression & Retrieved Expression \\
\hline
1 & $2\times 2$  & 4 & Similarity & Trace       & $A_{11} + A_{22}$           & $-0.677-A_{11} - A_{22}$ \\
2 &   $2\times 2$           & 4  &   Similarity                        & Determinant & $A_{11}A_{22} - A_{12}A_{21}$   & $A_{12}A_{21} - A_{11}A_{22}$ \\
\hline
3 & $3\times 3$  & 9 & Similarity & Trace       & $A_{11} + A_{22} + A_{33}$         & $A_{11} + A_{22} + A_{33}$ \\
4 & $3\times 3$ Antisymmetric & 9  &  Orthogonal                      & Sum of Principal Minors & $A_{12}^2 + A_{23}^2 + A_{13}^2$ & $A_{12}A_{21} + A_{23}A_{32} + A_{13}A_{31}$ \\
\hline
5 & $4\times 4$ & 16 & Similarity & Trace & $A_{11} + A_{22} + A_{33} + A_{44}$ &  $A_{11} + A_{22} + A_{33} + A_{44}$ \\
6 & $4\times4$ Field Strenth Tensor & 6 & Lorentz & Determinant & $E_1 B_1 + E_2 B_2 + E_3 B_3$ & $E_1 B_1 + E_2 B_2 + E_3 B_3$  \\
\hline
\end{tabular}}
\end{table}

\begin{table}[htbp]
\caption{Potentials}\label{table:tab_potentials}
\resizebox{\columnwidth}{!}{\begin{tabular}{|l|l|l|l|l|l|}
\hline
Experiment No. & $d$ & Potential $V$& Invariant & Analytical Expression & Retrieved Expression \\
\hline
7  & 2 & $\frac{1}{2}x^2$ & Energy       & $\frac{1}{2}v^2 + \frac{1}{2}x^2$ & $0.173/(v^2 + x^2)$   \\
8 &  2 & $\frac{1}{2}x^2 + \frac{1}{4}x^4$ &Energy & $\frac{1}{2}v^2 + \frac{1}{2}x^2 + \frac{1}{4}x^4$ & $v^2 + \exp(x^2)$ \\ 
9  &  2 & $\sin(x)$                         & Energy &  $\frac{1}{2}v^2 + \sin(x)$  & $-v^2 + \sin(x)/(-0.5) $ \\
10 &  2 & $\frac{1}{2}x^2 + \exp(x+1)$ & Energy & $\frac{1}{2}v^2 + \frac{1}{2}x^2 + \exp(x+1)$ & $v^2 + x^2 + \exp(x)/\sin(0.184)$ \\
\hline
11 & 4 & $-r^{-2}$ & Angular Momentum & $x_1 v_2 - x_2 v_1$ & $x_2 v_1 - x_1 v_2$ \\
\hline
\end{tabular}}
\end{table}

\begin{table}[htbp!]
\caption{Spacetime}\label{table:tab_spacetime}
\resizebox{\columnwidth}{!}{\begin{tabular}{|l|l|l|l|l|l|}
\hline
Experiment No. & $d$ & Transformation & Invariant & Analytical Expression & Retrieved Expression \\
\hline
12  & 4 & Lorentz Transformation & Spacetime Interval & $t^2 - x_1^2 - x_2^2 - x_3^2$ & $t^2 - x_1^2 - x_2^2 - x_3^2$  \\
\hline
\end{tabular}}
\end{table}

\label{sec:experiments}

\subsection{Results}

\begin{figure}[!h]
    \centering
    \includegraphics[width=0.9\textwidth]{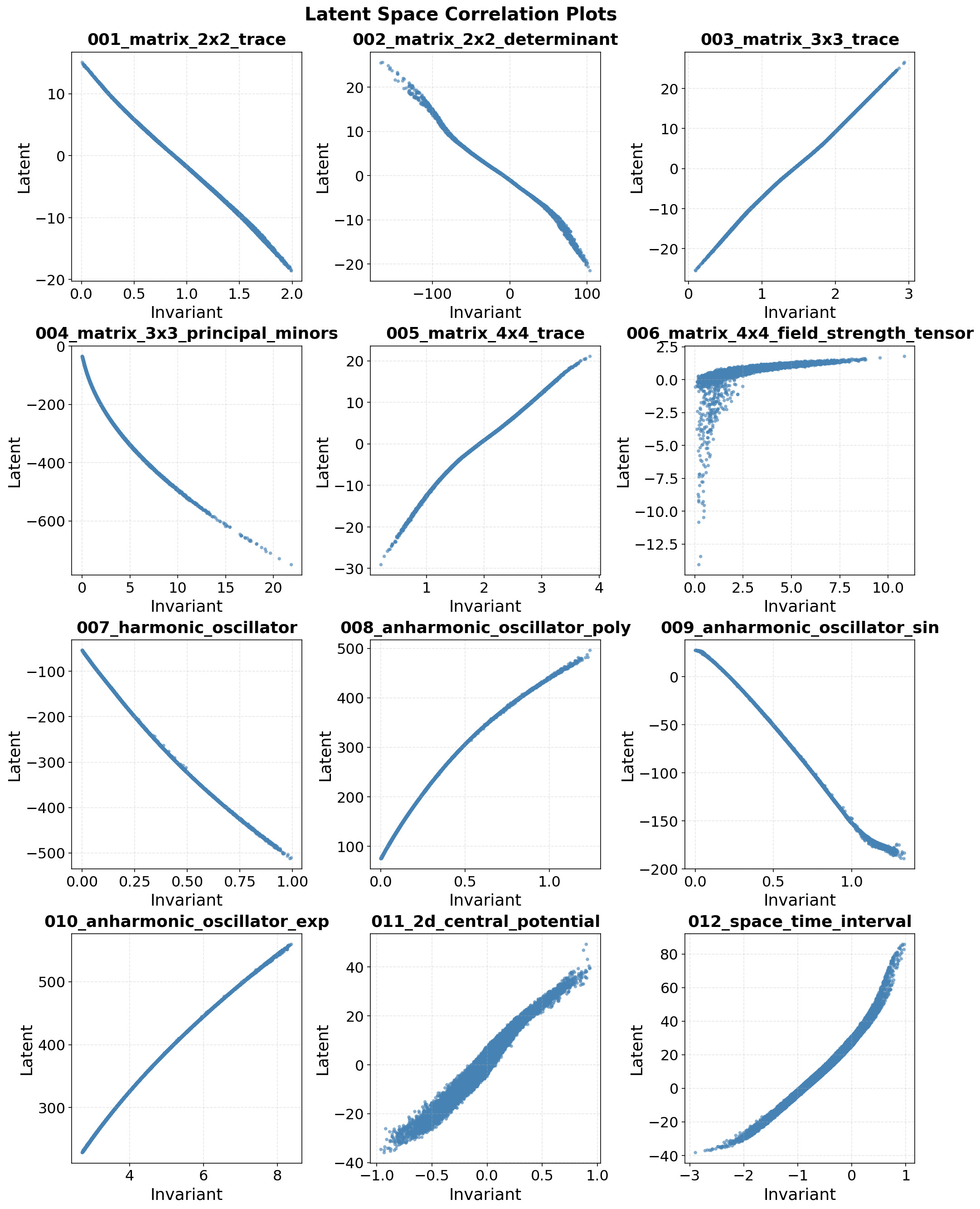}
    \caption{The latent space encodings of Siamese neural network applied to different data sets are compared with the corresponding ground truth concept for each data point. In all cases, it is possible to see a clear correlation. However, this correlation is mostly non-linear causing direct symbolic regression methods to fail, since they would attempt to fit additional variables for slopes and intercepts as well as the deformation to a non-linear dependency.}
    \label{fig:correlation}
\end{figure}

\begin{figure}[!h]
    \centering
    \includegraphics[width=\textwidth]{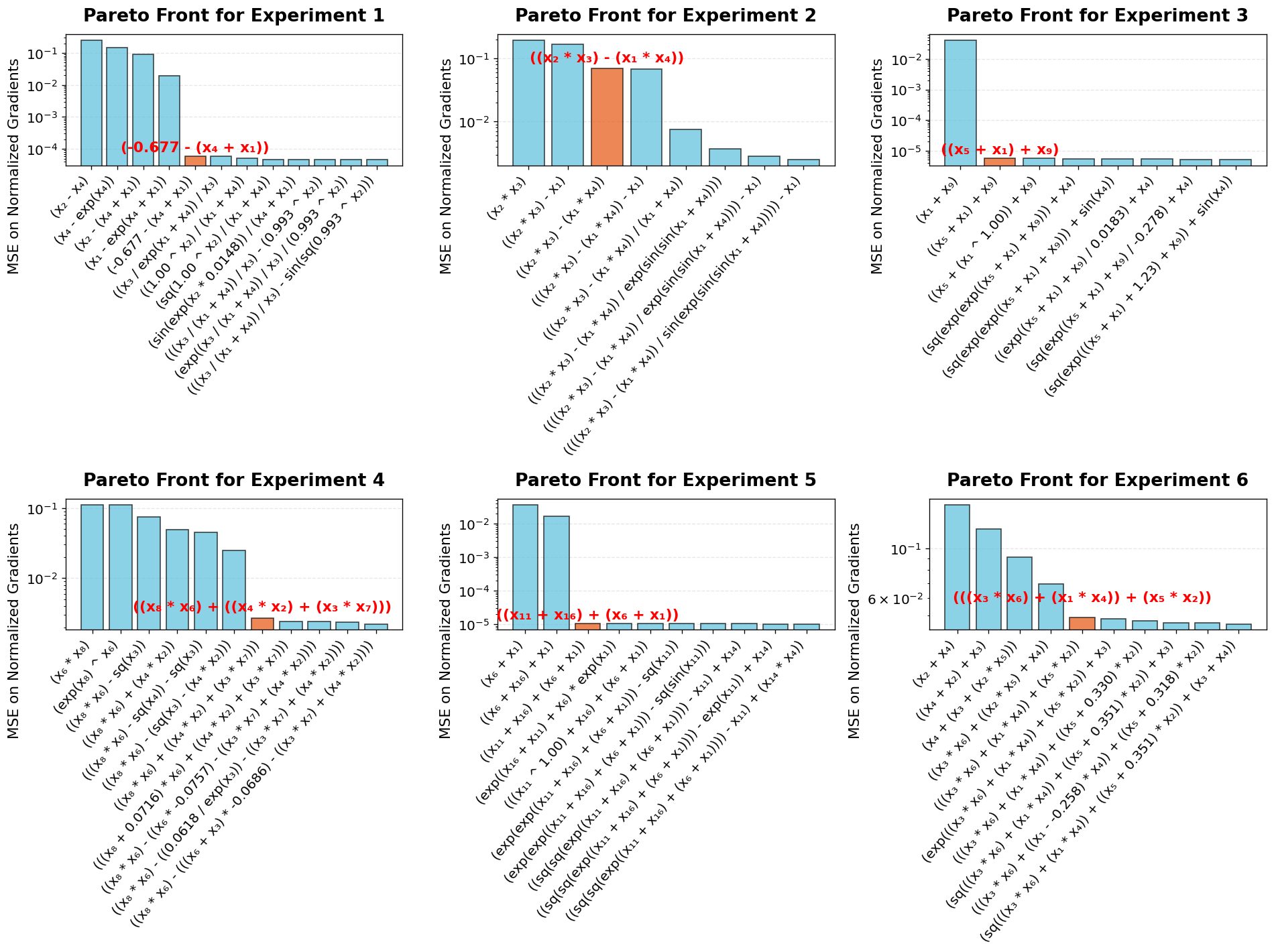}
    \caption{The Pareto front for experiments involving matrices, summarizing the results of the symbolic gradient-based interpretation framework to find a candidate concept that is contained in the corresponding neural network latent space. Several possible equations are plotted in order of decreasing Mean Square Error (MSE) and increasing complexity. The red bar indicates the candidate that resembles the ground truth concept, which is often found at the point of steepest change of the Pareto front.}
    \label{fig:pareto_conserved}
\end{figure}

\begin{figure}[!h]
    \centering
    \includegraphics[width=\textwidth]{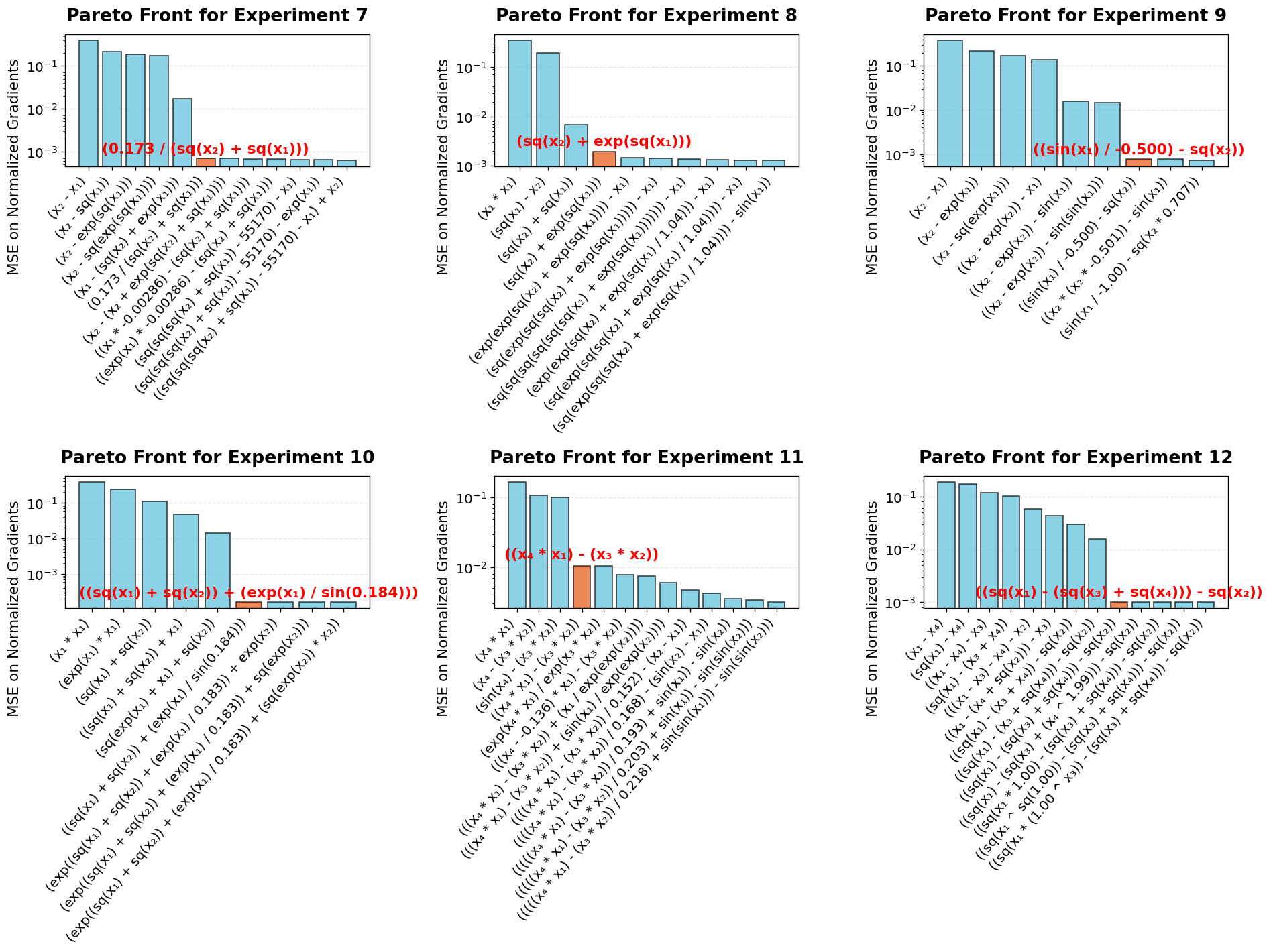}
    \caption{The Pareto front for experiments involving conserved quantities, summarizing the results of the symbolic gradient-based interpretation framework to find a candidate concept that is contained in the corresponding neural network latent space. Several possible equations are plotted in order of decreasing Mean Square Error (MSE) and increasing complexity. The red bar indicates the candidate that resembles the ground truth concept, which is often found at the point of steepest change of the Pareto front.}
    \label{fig:pareto_other}
\end{figure}

For each experiment, we use the method outlined in \autoref{sec:method} to obtain a set of predicted expressions from the symbolic search algorithm, which we present as a Pareto chart in \autoref{fig:pareto_other} and \autoref{fig:pareto_conserved}. The Pareto chart plots each of these expressions as a bar chart in decreasing order of loss and increasing complexity. Of these expressions, we identify the one that most closely matches the correct expression at the steepest drop of the Pareto front, translate the variables to the underlying experiments, and present it under the column titled \textit{retrieved expression} in tables \ref{table:tab_matrix}--\ref{table:tab_spacetime}. The correct expression is typically found after the steepest drop in the loss, corresponding to the lowest complexity symbolic solution that captures the ground truth. A notable exception to this rule arises when the network learns a good approximation to the desired expression, which can be rectified by increasing the sampling range used to produce the dataset. Since $H_g$ contains many different symbolic solutions that are all connected by an invertible transformation, it is possible to find different formulations of the ground truth at different complexity levels.

In our experiments, it was possible to retrieve all the correct ground truth expressions or, in one example, a very good approximation.  We also observe that the symbolic search algorithm might converge to a different formulation of the ground truth, by making simplifications to it or by expressing terms in a more complex manner. 

Of note is experiment 7, where the symbolic search algorithm refuses to simplify the retrieved expression $0.173/(v^2 + x^2)$ to $v^2 + x^2$. Further, in experiment 8, the resulting equation $v^2+\exp(x^2) \approx v^2 + 1 + x^2 + \frac{x^4}{2}$ is an approximation which matches the correct solution up to the fourth order in $x$. Since we initialized the initial conditions for this experiment with a mean of $0.5$ the 6th order is suppressed by $1/3! \times 0.5^6\approx0.0026$. We can confirm that by increasing the training data points to $200000$ the algorithm uncovers the correct ground truth. In experiment 10, $\sin(0.184)\approx 1/(2e)$ can be absorbed as prefactors and exponent leading to the correct discovery of the ground truth.

We compare the latent projection $f(\mathbf{X})$ for all inputs on the data set $\mathbf{X}$ to the true underlying concept $g(\mathbf{X})$. This can be visualized by plotting these quantities against each other in \autoref{fig:correlation}. Note that these correlation plots are not a necessary component of our interpretation framework and are solely used to highlight the non-linear manner in which the neural network encodes the concept. In most experiments, the values encoded in the latent spaces are highly correlated with the ground truth concepts. In fact, the correlation plots for the traces of the different matrices in \autoref{fig:correlation} are almost linear, which is expected as they can trivially be learned by a single-layer neural network with no non-linearities. In such cases, it is possible to use methods such as directly applying symbolic regression to the latent space to interpret the neural network. However, most invariants are significantly more complex, and the neural network will encode them in a non-linear manner, in which case interpretation by symbolic regression fails. All of these methods fail for the same reason - they attempt to retrieve the distorted version of the concept $\phi(g(\mathbf{x}))$, rather than the concept itself. In comparison, our method searches for a symbolic tree whose gradients are aligned with the network $f$. This means that the tree is not restricted to representing the distorted concept, and coupled with the complexity penalty of symbolic search, it yields the simplest possible expression whose gradients match the network $f$. We provide a comparison of our method to performing symbolic regression directly on the latent space \cite{cranmer2020discovering} in \autoref{appdx:direct_sr}, where only 7 of 12 experiments are successful. In our experiments, direct symbolic regression is capable of identifying invariants of polynomials up to second order. Of the 7 successful experiments, 3 of them are simply the trace, 2 are second order polynomials involving cross-terms, and the remaining two are also second order polynomials, but without any cross terms.  Interestingly, direct symbolic regression manages to retrieve a valid version of the ground truth expression for the sum of principal minors of $3\times 3$ antisymmetric matrices, which is encoded in a highly non-linear manner according to the correlation plot in \autoref{fig:correlation}. It is important to highlight that direct symbolic regression can rarely succeed in discovering the ground truth by accident even though it fails to approximate the latent projection. The reason behind this accidental success is likely based on finding a good solution on data-dense regions on the data manifold - see \autoref{fig:direct_sr_overlay} in \autoref{appdx:direct_sr}.

\section{Conclusions}
In this manuscript, we developed a framework to interpret any single neuron in neural network latent spaces in the form of a symbolic equation. It is based on employing symbolic search to find a symbolic tree that exhibits the same normalized gradients as the examined latent space neuron. The approach is suitable to interpret all kinds of neural networks applied to structured data within settings where concepts are formulated as scalar equations, like in science. The approach is limited by the expressibility of symbolic search algorithms and the challenge of isolating single neurons through bottlenecks or disentanglement.

We justify this procedure by defining an equivalence class of functions encoding the same concept, in which the membership criterion is that all members have parallel gradients at every point on the data manifold. Through this procedure, we enable the extraction of concepts encoded by latent space models.

We demonstrate the power of our approach by interpreting Siamese networks tasked with discovering invariants of matrices and conserved quantities of dynamical systems. We are able to uncover the correct equations or very good approximations in all of our examples. It is important to note that the symbolic search algorithm sometimes made clever approximations. For example, the anharmonic potential was summarized by an exponential function whose Taylor expansion agrees to fourth order in $x$. 

It is impossible to compare our results to other methods because our approach is the only general method that allows for the extraction of concepts encoded in latent spaces in closed form. As we have seen, sometimes the latent space encodings are approximately linearly correlated with the human-readable ground truth concept. In these cases, it is possible to retrieve the expression with traditional symbolic or polynomial regression. However, this is not the general case. It is important to note that there might exist publication bias towards linear encodings, where direct symbolic or polynomial regression methods accidentally succeed.

The pathways to scientific understanding via interpretable machine learning might lead down different roads. On one side there are inherently interpretable ML models, like PCA or support vector machines. On the other side, there are powerful artificial neural networks, which are difficult to interpret. Further, there is a middle ground implementing layers resembling symbolic calculations inside artificial neural networks. Until recently, none of these approaches was able to truly discover human-readable concepts from latent space models. We hope that through our approach many scientists will understand the potential discoveries that their latent space models might make.

The code to create all data and perform the experiments in this work can be found at \url{https://github.com/sjwetzel/PublicSymbolicGradientLatentSpaceInterpretation}.

\section{Acknowledgements}
We thank the National Research Council of Canada for their partnership with Perimeter on the PIQuIL. Research at Perimeter Institute is supported in part by the Government of Canada through the Department of Innovation, Science and Economic Development Canada and by the Province of Ontario through the Ministry of Economic Development, Job Creation and Trade. This work was supported by Mitacs through the Mitacs Accelerate program.

\section{Contributions}
SJW had the idea for the project and designed the research agenda. SJW provided the mathematical foundations, SJW and ZP were involved in writing code supporting this work. SJW and ZP wrote the manuscript and prepared figures.

\bibliographystyle{ACM-Reference-Format}

\newpage

\phantomsection
\addcontentsline{toc}{chapter}{Bibliography}
\bibliography{main}

\appendix
\section{Appendix}

\phantomsection
\appendix

\section{Equivalence of Equivalence Classes}
\subsection{Proof}
\label{sec:proof}

Let $g(\mathbf{x}),f(\mathbf{x}) \in C^1(D \subset \mathbb{R}^n,\mathbb{R})$ be continuously differentiable functions ($C^1(D\subset\mathbb{R}^n,\mathbb{R})$ is the vector space of differentiable functions from $D$ to $\mathbb{R}$) and $D\subset\mathbb{R}^n$ be the data manifold which is required to be compact and simply connected. Then $\tilde H_g= H_g=H_{g+}\cup H_{g-}$ where 
\begin{align}
\tilde H_g=\left\{ f(\mathbf{x}) \in C^1(D \subset \mathbb{R}^n,\mathbb{R}) \ | \exists \text{ invertible } \phi \in C^1(\mathbb{R},\mathbb{R}) \ : \ f(\mathbf{x}) = \phi(g(\mathbf{x}))\right\} 
\end{align}
and
\begin{align}
H_{g\pm}=\left\{ f(\mathbf{x}) \in C^1(D \subset \mathbb{R}^n,\mathbb{R}) \ | \ \frac{\nabla f(\mathbf{x})}{\lVert \nabla f(\mathbf{x})\rVert} = \frac{\pm\nabla g(\mathbf{x}) }{\lVert \nabla g(\mathbf{x})\rVert} \lor \nabla f(\mathbf{x})=\nabla g(\mathbf{x})=0, \ \forall \mathbf{x} \in D \right\} 
\end{align}

Proof: One can see that for each function $f\in \tilde H_g \ \phi:\nabla f(\mathbf{x}) = \phi'(g(\mathbf{x})) \nabla g(\mathbf{x})$ , hence the gradients are parallel and thus $\tilde H_g \subset H_g$.

It remains to be shown that for each function $f\in H_g \ \exists \phi : f(\mathbf{x})=\phi(g(\mathbf{x}))$. Let us focus on $f\in H_{g+}$, the proof is analogous for $H_{g-}$. Let us explicitly construct the function $\phi$ that maps between $f$ and $g$. Defining $\phi'$ through 
\begin{align}
\nabla f(\mathbf{x}) = \phi'(g(\mathbf{x})) \nabla g(\mathbf{x})\label{eq:grad}
\end{align}
omits the avoids the necessity of defining $\phi$ at locations where the gradients are zero. This definition leads to an integrable $\phi'(g(\mathbf{x})) \nabla g(\mathbf{x}) = \nabla f(\mathbf{x})$ because a) the images of $f(D)$ and $g(D)$ are compact, thus $\phi'$ maps between compact subsets of $\mathbb{R}$ and b) $\phi'$ is continuous. For any simply connected $D \subset \mathbb{R}^n$ we can define the $C^1$-curve $\mathbf{x}:[t_0,t_1]\rightarrow D$, thus a variable transformation within the calculation of the contour integral yields:

\begin{align}
\phi(g(\mathbf{x}(t_1)))-\phi(g(\mathbf{x}(t_0)))&=\int_{g(\mathbf{x}(t_0))}^{g(\mathbf{x}(t_1))} \phi'(g)\ dg \\
&=\int_{\mathbf{x}(t_0)}^{\mathbf{x}(t_1)} \phi'(g(\mathbf{x})) \nabla g(\mathbf{x})\cdot d\mathbf{x} \\
&=\int_{t_0}^{t_1} \phi'(g(\mathbf{x}(t))) \nabla g(\mathbf{x}(t)) \cdot \dot{ \mathbf{x}}(t) \ dt \\
&\stackrel{eq.\ref{eq:grad}}{=}\int_{t_0}^{t_1}  \nabla f(\mathbf{x}(t))  \cdot \dot{ \mathbf{x}}(t) \ dt \\
&=\int_{x(t_0)}^{x(t_1)}  \nabla f(\mathbf{x})  \ d \mathbf{x} \\
&=\int_{f(x(t_0))}^{f(x(t_1))}    \ d f \\
&=f(\mathbf{x}(t_1)) - f(\mathbf{x}(t_0))
\end{align}
Similarly, one can proof the existence of $\tilde \phi:\tilde \phi(f(\mathbf{x}))=g(\mathbf{x})$ such that $f(\mathbf{x})=\phi(\tilde\phi(f(\mathbf{x})))$ and thus $\phi$ is invertible. While this proof assumes $f\in H_{g+}$ it is analogously possible to construct $\phi$ for $f\in H_{g-}$. Having explicitly constructed $\phi$ proofs $H_g=H_{g+}\cup H_{g-}=\tilde H_g$.

\subsubsection{Assumptions}
\label{sec:assumptions}

In a practical machine learning applications not all assumptions from the prior section that ensure $H_g = \tilde H_g$ hold true. However, even then $H_g \approx \tilde H_g$ can provide a good approximation that allows for a retrieval of the function that a neuron encodes.

A machine learning data set approximate the data manifold $D \subset \mathbb{R}^n$. If there is a divergence in the function that the machine learning model is supposed to approximate, the data set might not be closed and thus not compact. A data manifold $D$ might not be simply connected, especially if it is in the form of categorical data or images.

A neural network classifier, if successfully trained, tends to approximate a categorical output, which is neither continuous nor differentiable. However, this binary output is typically an approximation mediated by sigmoid or softmax activation functions, which indeed are continuously differentiable. Still, interpreting artificial neural networks with the framework introduced in this paper experiences numerical artifacts if a gradient is taken from a network that contains sigmoid or softmax activation functions. For this reason, I suggest avoiding these activation functions in the design of hidden layers and removing them from the output neuron during the interpretation process (the same argument holds true for tanh or related activation functions).

The above definitions of equivalence classes could be extended to piecewise $C^1(\mathbb{R}^n,\mathbb{R})$ functions. This function set contains many artificial neural networks that include piecewise differentiable activation functions like $\text{ReLU}(x)=\max(0,x)$. However, this causes problems when evaluating derivatives close to $\text{ReLU }(x)=0$. In practice, one can observe that piecewise $C^1$ activation functions lead to computational artifacts when calculating gradients. Hence, I suggest using $\text{ELU}=\exp(x)-1 |x\leq 0, x | x > 0$ as the preferred activation function in hidden layers.

\section{Preliminaries}
\label{sec:preliminaries}

\subsection{Matrix Invariants}
\subsubsection{Invariants of Rank-Two Tensors Under Similarity Transformations}

For any $n\times n$ matrix $A$, we can compute the similarity transform $B=CAC^{-1}$, where $C$ is any invertible matrix. Using the cyclic property of trace, 
\begin{align*}
    \text{tr}(B) = \text{tr}(CAC^{-1}) = \text{tr}(AC^{-1}M) = \text{tr}(A).
\end{align*}
Furthermore, 
\begin{align*}
    \det(B) = \det(CAC^{-1}) = \det(C)\det(A)\frac{1}{\det(C)} = \det(A).
\end{align*}
Hence, both the trace and determinant are invariant under this basis change. It is straightforward to see that the following expression, called the \textit{sum of principle minors}, is also basis-invariant:
\begin{align*}  
    \text{tr}(\text{tr}(A)^2 - \text{tr}(A^2)).
\end{align*}
Together, these three comprise the principal invariants of rank-two tensors:
\begin{align}    
    & I_1 = \text{tr}(A), \\
    \label{eq:sum_principal_minor}
    & I_2 = \text{tr}(\text{tr}(A)^2 - \text{tr}(A^2)), \\
    & I_3 = \det(A).
\end{align}
In this case, the placeholder operator $M$ is $M(A)=CAC^{-1}$.
\subsubsection{$3 \times 3$ Antisymmetric Matrices}
\label{sec:antisymmetric_matrices}
In the case of antisymmetric $n\times n$ matrices of odd size, the number of principal invariants reduces to one. The trace of an antisymmetric matrix is $0$, so $I_1=\text{tr}(A)=0$, and any antisymmetric square matrix of odd size $n$ must have at least one zero-eigenvalue, so $I_3=\det(A)=0$. 
We treat the case of a $3\times 3$ antisymmetric matrix, in which case $I_2$ can be written in terms of its entries as,
\begin{align*}    
    I_2 = A_{11}A_{22} + A_{22}A_{33} + A_{11}A_{33} - A_{12}A_{21} - A_{23}A_{32} - A_{13}A_{31}.
\end{align*}
Since the diagonal elements of $A$ are $0$ and $A_{ij}=-A_{ji}$, the expression for $I_2$ is simplified:
\begin{align}    
    \label{eq:antisymmetric_sum_principal_minor}
    I_2 = A_{12}^2 + A_{23}^2 + A_{13}^2.
\end{align}
For the case of antisymmetric matrices it is important to restrict the transformation $M(A)=CAC^{-1}$ to be mediated by orthogonal matrices $C\in O(3)$ to preserve antisymmetricity.

\subsubsection{Invariants of the Field Strength Tensor Under the Lorentz Transformation}
\label{sec:EM}

The electromagnetic field strength tensor unifies the magnetic $\mathbf{B}$ and electric fields $\mathbf{E}$

\begin{align}
F^{\mu\nu} = \begin{bmatrix}
     0     & -E_x/c & -E_y/c & -E_z/c \\
     E_x/c &  0     & -B_z   &  B_y    \\
     E_y/c &  B_z   &  0     & -B_x   \\
     E_z/c & -B_y   &  B_x   &  0
  \end{bmatrix}
\end{align}
where $c=1$ is the speed of light. The tensor \( F_{\mu\nu} \) is antisymmetric, meaning \( F_{\mu\nu} = -F_{\nu\mu} \). Under Lorentz transformations, the invariants of the electromagnetic field strength tensor \( F_{\mu\nu} \) are preserved. Its invariants include the scalar \( \mathbf{B} \cdot \mathbf{E} \) and the quantity \( \frac{1}{2} F_{\mu\nu} F^{\mu\nu} \). Specifically,
\[
\mathbf{B} \cdot \mathbf{E} = \det(F_{\mu\nu})
\]
and
\[
\left| \mathbf{B} \right|^2 - \left| \mathbf{E} \right|^2 = \frac{1}{2} F_{\mu\nu} F^{\mu\nu}.
\]
These invariants highlight the consistency of electromagnetic properties across different inertial frames. The placeholder operator implements $M(A)=\Lambda A \Lambda^\top$, where $\Lambda$ is the Lorentz transformation.

\subsection{Space-Time Interval}
\label{sec:spacetime}
In special relativity, Minkowski spacetime is a four-dimensional continuum that combines three spatial dimensions with one time dimension. This framework allows for a unified description of space and time, where events are described by four coordinates $
(t,x,y,z)$, and the separation between events is invariant under Lorentz transformations. The distance between events in this spacetime is determined by the Minkowski metric, which is defined by the metric tensor,
\begin{align*}
    \eta_{\mu \nu} = \text{diag}(-1, 1, 1, 1)
\end{align*}
This metric defines a scalar product for any two four-vectors \(x\) and \(y\) as,
\begin{align*}
    \langle x, y \rangle = \eta_{\mu \nu} x^\mu y^\nu = x^\mu y_\mu
\end{align*}
where \(x^\mu\) and \(y^\mu\) are the components of the four-vectors \(x\) and \(y\). The spacetime interval \(s\), which remains constant under Lorentz transformations, is given by,
\begin{align*}
    \langle x, x \rangle = -t^2 + x^2 + y^2 + z^2 = s^2
\end{align*}

The Lorentz group, which consists of transformations that preserve this scalar product in Minkowski spacetime, is denoted as,
\begin{align*}
    O(3,1) = \left\{ \Lambda \in M(\mathbb{R}^4) \mid \langle \Lambda x, \Lambda y \rangle = \langle x, y \rangle, \forall x, y \in \mathbb{R}^4 \right\}
\end{align*}

In this case, the placeholder operator performs $M(A)=\Lambda A$, where $\Lambda$ is the Lorentz transformation.

\subsection{Dynamical Systems}
\label{sec:dynamical_systems}
In dynamical systems involving motion in a potential, conservation principles are fundamental. In one-dimensional systems, energy is invariant, meaning the total energy—comprising both kinetic and potential components—remains constant in an isolated system. In two-dimensional systems, both energy and angular momentum are conserved, provided the potential is central (i.e., depends only on the radial distance). These invariants are crucial for understanding and modeling dynamical behaviors in both 1D and 2D contexts. The operator \( M \) evolves the system by mapping a state vector \( \mathbf{x} \) to its time-evolved counterpart, \( M(\mathbf{x}) \), representing the system's state at a later time.

\section{Implementation Details}
\subsection{Symbolic Regression Hyperparameters}
\label{appdx:sr_params}
To reproduce this experiment, the \textbf{SymbolicRegression.jl} library was used to perform symbolic search with a custom loss function targeting gradient alignment. The model was configured with binary operators (\texttt{+, -, *, /, $\hat{}$, div}) and unary operators (\texttt{sqrt, square, sin, exp}). Complexity penalties were assigned as follows: constants had a complexity of \texttt{3}, while operators had complexities of \texttt{sqrt => 4}, \texttt{square => 4}, \texttt{sin => 5}, and \texttt{exp => 5}. The training process involved \texttt{niterations=200}, \texttt{batch\_size=25}, \texttt{early\_stop\_condition=1e-10}, and a \texttt{maxsize=25} constraint on the equation size. Simplification of equations, optimization of constants, and automatic differentiation were enabled to improve the accuracy and interpretability of the resulting expressions.

To reproduce this experiment, the \textbf{SymbolicRegression.jl} library was used to perform symbolic search with a custom loss function targeting gradient alignment. The model was configured with binary operators (\texttt{+, -, *, /, $\hat{}$, div}) and unary operators (\texttt{sqrt, square, sin, exp}). Complexity penalties were assigned as follows: constants had a complexity of \texttt{3}, while operators had complexities of \texttt{sqrt => 4}, \texttt{square => 4}, \texttt{sin => 5}, and \texttt{exp => 5}. The training process involved \texttt{niterations=200}, \texttt{batch\_size=25}, \texttt{early\_stop\_condition=1e-10}, and a \texttt{maxsize=25} constraint on the equation size. Simplification of equations, optimization of constants, and automatic differentiation were enabled to improve the accuracy and interpretability of the resulting expressions.

\textcolor{red}{}

\subsection{Training Hyperparameters}
\label{appdx:neural_network}
All experiments use the Adam optimizer and the scheduler class \texttt{ReduceLROnPlateau} from the PyTorch library. For all experiments, the sub-network $f$ is a fully-connected feedforward network designed as follows:

\begin{itemize}
    \item An input layer 
    \item Two hidden layers
    \item An output layer with a single neuron
\end{itemize}

The layer sizes for each experiment are given in \autoref{table:network_params}. A ReLU activation is used at the output of each neuron, except for the final one. Furthermore, the margin for the triplet loss is $\alpha=1$.

\begin{figure}[!h]
    \label{fig:neural_network}
    \centering
    \includegraphics[width=0.8\textwidth]{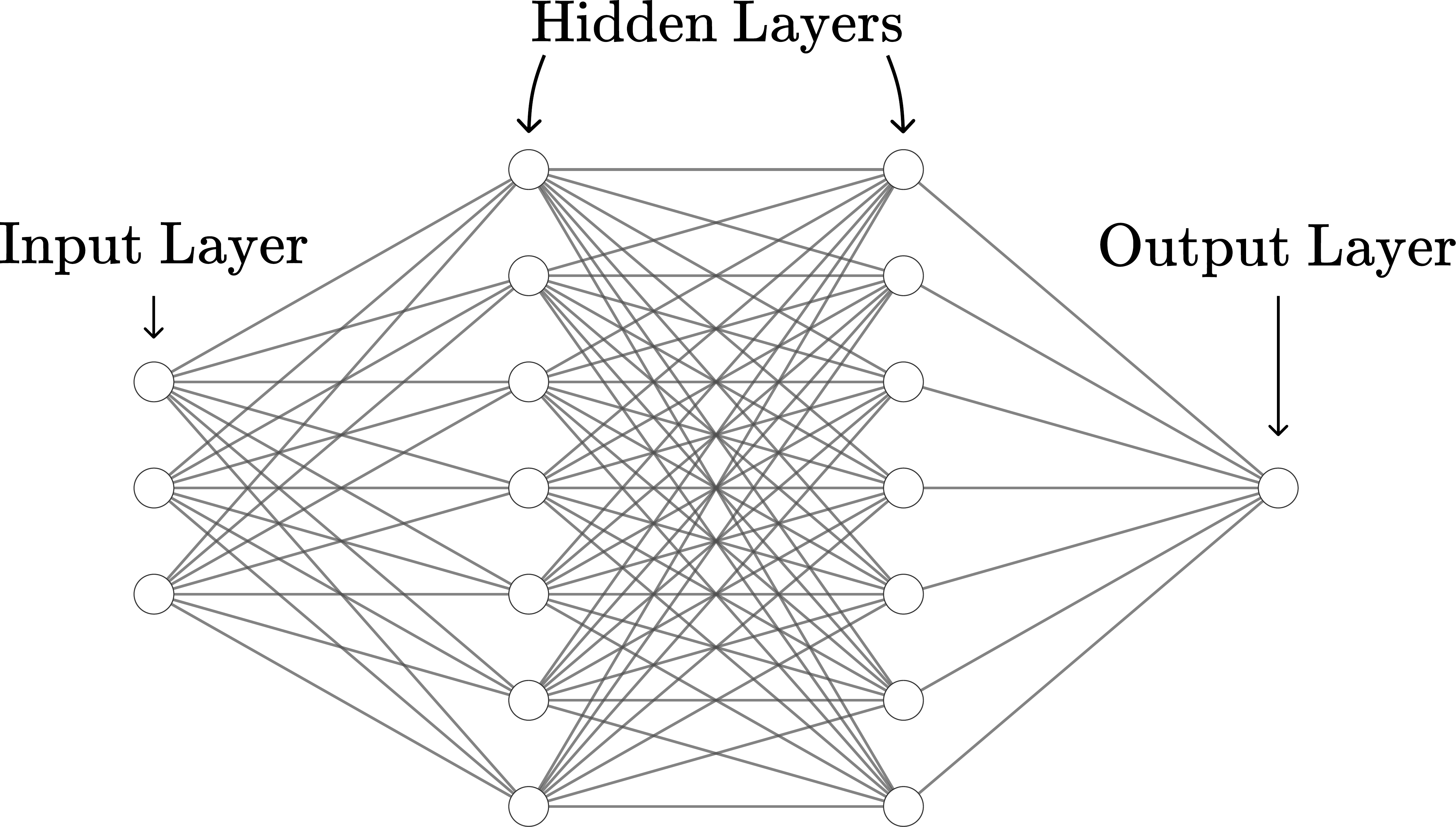}
    \caption{A visual depiction of the architecture we use for each sub-network $f$. The input layer has the same number of neurons as the dimensionality of the input. We use two hidden layers with, followed by an output layer with a single neuron. This final neuron is the latent space neuron that we interpret in our examples.}
\end{figure}

\begin{table}[htbp]
\caption{Training Hyperparameters}\label{table:network_params}
\resizebox{\columnwidth}{!}{\begin{tabular}{|l|l|l|l|l|l|l|l|l|l|}
\hline
Exp. No. & Learning Rate & Weight Decay & Batch Size & Input Size & Hidden Size & Output Size & Epochs & Factor & Patience \\
\hline
1 & 0.001 & 0.0001 & 256 & 4 & 256 & 1 & 100 & 0.2 & 10 \\
2 & 0.001 & 0.0001 & 256 & 4 & 256 & 1 & 300 & 0.2 & 10 \\
3 & 0.001 & 0.0001 & 256 & 9 & 256 & 1 & 100 & 0.2 & 10 \\
4 & 0.001 & 0.0001 & 256 & 9 & 256 & 1 & 100 & 0.2 & 10 \\
5 & 0.001 & 0.0001 & 256 & 16 & 256 & 1 & 300 & 0.2 & 10 \\
6 & 0.001 & 0.0001 & 256 & 6 & 256 & 1 & 300 & 0.5 & 10 \\
7 & 0.001 & 0.0001 & 256 & 2 & 256 & 1 & 300 & 0.5 & 10 \\
8 & 0.001 & 0.0001 & 256 & 2 & 256 & 1 & 300 & 0.5 & 10 \\
9 & 0.001 & 0.0001 & 256 & 2 & 256 & 1 & 300 & 0.5 & 10 \\
10 & 0.001 & 0.0001 & 256 & 2 & 256 & 1 & 300 & 0.5 & 10 \\
11 & 0.001 & 0.0001 & 256 & 4 & 256 & 1 & 300 & 0.5 & 10 \\
12 & 0.001 & 0.0001 & 256 & 4 & 256 & 1 & 300 & 0.5 & 10 \\
\hline
\end{tabular}}
\end{table}

\section{Dataset Generation}
\label{appdx:dataset}
We retrieve the invariants of matrices and various physical systems using our method. We consider invariants of matrices under similarity and Lorentz transformations. Additionally, we investigate dynamical systems characterized by a variety of potentials, as well as the invariants in Minkowski spacetime. We choose the mass $m=1$ and spring constant $k=1$ where applicable.

\subsection{Datasets and Training}
\subsubsection{Invariants Under the Similarity Transformation}
\label{sec:matrix_dataset}
In experiments 1-3, 5 in Table \ref{table:tab_matrix}, we search for the trace and determinant of matrices under the similarity transformation. Each data point is a triplet consisting of three matrices of dimension $n$: an anchor matrix $A$, a positive example $P$, and a negative example $N$. The anchor is sampled by generating a random matrix. Each entry is sampled from a uniform distribution between $\left[\alpha, \beta\right]$. We try $\left[0, 1\right]$, and $\left[-4, 4\right]$. Neither choice affects the model's ability to learn the invariant.

The positive example shares one or more invariants with the anchor. In the case of the similarity transformation, these invariants are the trace and determinant. To this end, we sample a $n \times n$ invertible matrix $M$ and apply the similarity transformation $P=MAM^{-1}$. The negative example should not share invariants with the anchor, which is trivially achieved by sampling another matrix $N$, which is almost certainly characterized by different invariants.

In practice, we find that the neural network prefers to learn the trace. To discover a second invariant, such as the determinant, we sample triplets in which all matrices have the same trace. The network can no longer rely on the trace to identify similar matrices, or to distinguish between dissimilar ones, as the trace no longer provides any useful information for this task. Instead, an alternative invariant must be learned, which in this case is the determinant. This can be done for any number of invariants: upon discovery of the first one, it can be made constant across the entire dataset to force the neural network to learn another.

\subsubsection{Invariants of Antisymmetric Matrices}

For antisymmetric matrices in experiment 4, we prepare our dataset in the same way as we describe in \ref{sec:matrix_dataset}. We first sample an antisymmetric $3\times 3$ matrix for the anchor $A$, followed by a similarity transformation for the positive sample $P$. Finally, we sample a new antisymmetric matrix for the negative sample $N$. While both the anchor and negative samples are antisymmetric, the positive sample does not inherit this property under the transformation $P=MAM^{-1}$ when $M$ is not orthonormal, because antisymmetry is not preserved under a general change of basis. Hence, we use all 9 entries of the matrix as input, although we acknowledge that one could easily enforce that $M$ is orthonormal, in which case only 3 inputs would be needed from each of $A$, $P$, and $N$. 

Since we use the antisymmetric anchor matrix $A$ as input when computing $\nabla_\mathbf{x} f(\mathbf{x})$, we expect that the result of symbolic search would simplify to the invariant in \ref{sec:antisymmetric_matrices}, which is invariant under the similarity transformation. 

The entries of each matrix are sampled from a normal distribution with $\mu=0$ and $\sigma=1$.

\subsubsection{Invariants Under the Lorentz Transformation}
In experiment 6, we apply the Lorentz transformation to the field strength tensor $F_{\mu \nu}$, which gives rise to the Lorentz invariants in \ref{sec:EM}. Since the antisymmetry of $F_{\mu \nu}$ is preserved under the Lorentz transformation, each member of a triplet is antisymmetric, so we only use the 6 off-diagonal entries above (or equivalently below) the main diagonal as our input to the neural network. The anchor is a vector of these 6 entries from $F_{\mu \nu}$.

The entries of each matrix are sampled from a uniform distribution between $\left[0, 1\right]$. 

\subsubsection{Potentials}

The experiments in \autoref{table:tab_potentials} correspond to motion in a potential, where we simulate trajectories by randomly sampling initial positions and velocities, and subsequently evolve these systems according to Hamilton's equations. 
For each triplet $(x_A, x_P, x_N)$, the anchor $x_A$ and positive sample $x_P$ are measurements at two different points along the same trajectory, while the negative sample $x_N$ is sampled from a different trajectory. The network must determine whether or not two measurements belong to the same particle. See \ref{sec:dynamical_systems} for details regarding the invariants.

The dataset is generated with mass \( m = 1 \), spring constant \( k = 1 \), and a time grid \( t \in [0, 5] \) with 10,001 points. Initial conditions are sampled from \([0, 1]^2\), and trajectory points are selected using random indices \( i, j \) from the solution.


\subsubsection{Spacetime}
In experiment 12 in \autoref{table:tab_spacetime}, each triplet again consists of an anchor $x_A$, a positive sample $x_P$, and a negative sample $x_N$. The anchor is a randomly sampled four-vector representing an event in Minkowski spacetime. Each entry is sampled from a uniform distribution between $\left[0, 1\right]$. The positive sample is generated by applying a Lorentz transformation to the anchor, ensuring that the spacetime interval remains invariant. The negative sample, on the other hand, is another randomly generated four-vector that does not share the same spacetime interval as the anchor, allowing the neural network to distinguish between vectors that do and do not preserve this invariant.

\section{Results of Direct Symbolic Regression}
\label{appdx:direct_sr}
We compare our method to direct symbolic regression, which is can only recover the ground truth if the latent space encodes it in a linear manner. In this case, we obtain latents from the trained neural network, and supply them to the standard symbolic regression algorithm \cite{Cranmer2023}, which then attempts to extract a symbolic equation from the dataset consisting of the pairs $(X, \text{latents})$. In \autoref{table:direct_sr}, we 
apply direct symbolic regression on the same experiments as in tables 1-3. We report the retrieved expression only if the correct expression is retrieved. Notably, only 7 out of 12 experiments succeed with direct symbolic regression. We also note that in the symbolic gradient experiments, we report the first result. In symbolic regression we allowed for multiple runs, including up to $200000$ data points. It is interesting to see that in one case, symbolic regression could recover a ground truth equation while failing to approximate the latents \autoref{fig:correlation}.

\begin{table}[htbp]
\caption{Results of Direct Symbolic Regression}\label{table:direct_sr}
\begin{center}
  \begin{tabular}{|l|l|l|}
  \hline
  Exp. No. & Expression Retrieved? & Expression  \\
  \hline
  1  & Yes  & $\frac{A_{11} + A_{22}}{-0.12}$ \\
  2  & Yes  &  $A_{21}A_{12} - A_{22}A_{11}$ \\
  3  & Yes  &  $A_{11} + A_{22} + A_{33} - 1.489$ \\
  4  & Yes  &  $3.239 - A_{12} A_{12} - A_{23} A_{23} - A_{13} A_{13}$ \\
  5  & Yes  & $A_{11} + A_{22} + A_{33} + A_{44} - 1.907$  \\
  6  & No   & - \\
  7  & Yes  & $394.111 x^2 - (-399.431) v^2$ \\
  8  & No   & - \\
  9  & No   & - \\
  10 & No   & - \\
  11 & Yes  & $x_2 v_1 - x_1 v_2$ \\
  12 & No   & - \\
  \hline
  \end{tabular}
\end{center}
\end{table}

\begin{figure}
    \centering
    \includegraphics[width=0.9\linewidth]{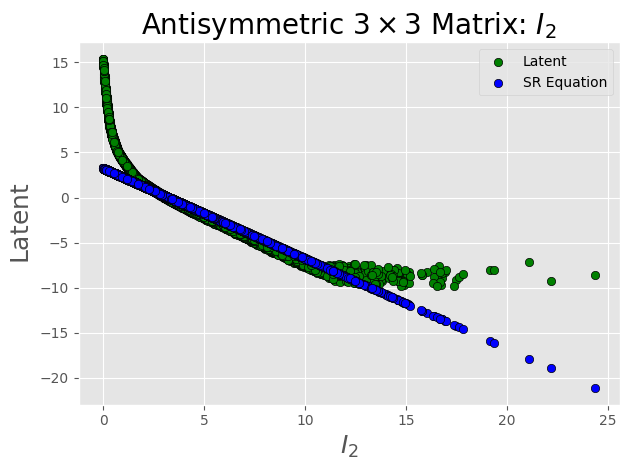}
    \caption{Direct symbolic regression on the latents typically fails when the concept is encoded in a non-linear manner. However, in the peculiar case of experiment 4 in \autoref{table:direct_sr}, this method retrieves the ground truth expression. This is likely because it is fitting to a data-dense region of the non-linear plot, as shown in this figure. The blue line represents the equation extracted through direct symbolic regression. It almost perfectly passes through a linear sub-region of the correlation curve.}
    \label{fig:direct_sr_overlay}
\end{figure}

\end{document}